\documentclass{article}

\usepackage[hidelinks]{hyperref}
\usepackage{PRIMEarxiv}
\usepackage{orcidlink}
\usepackage{algpseudocode}
\usepackage{algorithm}
\usepackage{amsmath}
\usepackage{float}
\usepackage[utf8]{inputenc} 
\usepackage[T1]{fontenc}    
\usepackage{hyperref}       
\usepackage{url}            
\usepackage{booktabs}       
\usepackage{amsfonts}       
\usepackage{fancyhdr}       
\usepackage{graphicx}       
\usepackage{authblk}

\title{Potrika: Raw and Balanced Newspaper Datasets in the Bangla Language with Eight Topics and Five Attributes
}

\author[1]{Istiak Ahmad\orcidlink{0000-0001-9914-4116}}
\author[1]{Fahad AlQurashi\orcidlink{0000-0002-7919-747X}}
\author[2,*]{Rashid Mehmood\orcidlink{0000-0002-4997-5322}}

\affil[1]{Department of Computer Science, Faculty of Computing and Information Technology, King Abdulaziz University, Jeddah 21589, Saudi Arabia}

\affil[2]{High Performance Computing Center, King Abdulaziz University, Jeddah 21589, Saudi Arabia}

\affil[*]{Corresponding author: RMehmood@kau.edu.sa}

\begin{document}
\maketitle

\begin{abstract}
Knowledge is central to human and scientific developments. Natural Language Processing (NLP) allows automated analysis and creation of knowledge. Data is a crucial NLP and machine learning ingredient.
The scarcity of open datasets is a well-known problem in machine and deep learning research. This is very much the case for textual NLP datasets in English and other major world languages. For the Bangla language, the situation is even more challenging and the number of large datasets for NLP research is practically nil.
We hereby present Potrika, a large single-label Bangla news article textual dataset curated for NLP research from six popular online news portals in Bangladesh (Jugantor, Jaijaidin, Ittefaq, Kaler Kontho, Inqilab, and Somoyer Alo) for the period 2014-2020. The articles are classified into eight distinct categories (National, Sports, International, Entertainment, Economy, Education, Politics, and Science \& Technology) providing five attributes (News Article, Category, Headline, Publication Date, and Newspaper Source).  
The raw dataset contains 185.51 million words and 12.57 million sentences contained in 664,880 news articles. Moreover, using NLP augmentation techniques, we create from the raw (unbalanced) dataset another (balanced) dataset comprising 320,000 news articles with 40,000 articles in each of the eight news categories. Potrika contains both the datasets (raw and balanced) to suit a wide range of NLP research. By far, to the best of our knowledge, Potrika is the largest and the most extensive dataset for news classification.
\end{abstract}

\keywords{Natural Language Processing\and Single-Label Text Dataset\and Bangla Text Analysis\and News Article Analysis\and Machine Learning\and Deep Learning}

\section{Introduction}

Knowledge is central to human and scientific developments. Natural Language Processing (NLP) allows automated analysis and creation of knowledge. Data is a crucial NLP and machine learning ingredient. 
Textual datasets are critically needed for NLP and machine learning research and development, in all human languages including English, more so in languages such as Bangla where NLP research is limited. 
The existing datasets for newspaper classification in the Bangla language are limited in their size (the number of news article, sentences, and words) and attributes (see Section~\ref{sec:description} where we provide a summary of the existing datasets and establish the research gap). 
	
We present here Potrika, the largest and the most extensive dataset to date for news classification containing 185.51 million words and 12.57 million sentences in 664,880 news articles, providing five attributes, classified into eight distinct categories. The data contains news articles for the period 2014-2020. It is a single-label dataset that can be used for text classification, text summarization, text generation, and other NLP-related research. It is made available publicly and freely. 
	
Potrika comprises two datasets, a raw dataset, and a balanced dataset created using NLP augmentation techniques. This balanced dataset contains 320,000 news articles with 40,000 articles in each of the eight news categories. 
Potrika can be used as a benchmark dataset. It may, for example, be used to create a diversity of word embedding models. Aside from that, Potrika comprises eight distinct news categories that have been carefully picked to reduce ambiguity and, as a consequence, render it robust for text classification. It may also be used to summarize content by referring to the news article and its associated headline. Furthermore, it can be used to investigate NLP techniques focussed on temporal aspects, such as identifying which category has the most influence at a particular time or which keywords are utilized more frequently over time.
	
This extensive dataset, Potrika, provides an opportunity to investigate and improve the efficacy of applying machine learning and deep learning algorithms for natural language processing on textual data in the Bangla language. The Potrika datasets (the raw dataset and the balanced dataset) are described in detail along with various dataset statistics, examples of news article with associated categories, and data collection method, architecture, and algorithms. 

The rest of the paper is organised as follows. Section~\ref{sec:description} provides a description of the Potrika dataset and its comparison with the existing datasets. Section~\ref{sec.design} provides the design and methodology of data collection and Section~\ref{sec.conclusion} concludes the paper.

\begin{table}[!htb]
	\centering
	\caption{Potrika: Comparison with Existing Bangla Newspaper Datasets}
	\label{tab.relatedResearch}
	\begin{tabular}{ccccclc} 
		\hline
		Dataset & \begin{tabular}[c]{@{}c@{}} Articles\\ (K) \end{tabular} & \begin{tabular}[c]{@{}c@{}} Sentences\\(M) \end{tabular} & \begin{tabular}[c]{@{}c@{}} Words\\(M) \end{tabular}  & \begin{tabular}[c]{@{}c@{}} Categories\\(Number) \end{tabular}   & Categories   &   Balanced                                                                                                                                                                                     \\ 
		\hline
		\begin{tabular}[c]{@{}c@{}} BARD~\cite{alam2018bard}\\2018\end{tabular}   & 376 & --        & 88 & 5           & \begin{tabular}[l]{@{}l@{}}Economy, State \\Entertainment,\\ International,\\ Sports,\end{tabular}  & $\times$                                                                                                   \\ 
		\hline
		\begin{tabular}[c]{@{}c@{}} Hossan~\cite{zakir40k}\\2018\end{tabular}  & 40    & --        & --          & 12 & \begin{tabular}[l]{@{}l@{}}Economy, \\Opinion,\\Entertainment, \\International, \\Sports, \\Bangladesh,\\ Technology,\\Education, \\Durporobash\\Art-and-literature, \\North America,\\Lifestyle\end{tabular} & $\times$ \\ 
		\hline
		\begin{tabular}[c]{@{}c@{}} Shahin~\cite{shahin2020classification}\\2020\end{tabular}  & 10    & --        & --          & 8           & \begin{tabular}[l]{@{}l@{}}Economy, \\International,\\Entertainment, \\Sports,\\ Science \& \\ Technology,\\Opinion, Politics \\Country\end{tabular} & $\times$                                                          \\ 
		\hline
		\hline
		\begin{tabular}[c]{@{}c@{}}\textbf{Potrika}~\cite{istiakdata2021}\\2021\end{tabular} & 665   & 125 & 185 & 8           & \begin{tabular}[l]{@{}l@{}}Economy, \\Entertainment, \\International, \\Sports,\\National,~Politics\\Science \& \\Technology, \\Education\end{tabular}         & \checkmark                                                \\
		\hline
	\end{tabular}
\end{table}

\section{Data Description}\label{sec:description}

Data is a crucial NLP and machine learning ingredient. NLP allows automated analysis and creation of knowledge -- knowledge that is central to human and scientific developments. This paper proposes the Potrika dataset that is a vast collection of news articles in the Bangla language collected from six prominent news portals in Bangladesh, including Jugantor \cite{Jugantor}, Jaijaidin \cite{Jaijaidin}, Ittefaq \cite{Ittefaq}, Kaler Kontho \cite{KalerKontho}, Inqilab \cite{Inqilab}, and Somoyer Alo \cite{SomoyerAlo}. 

The scarcity of open datasets is a well-known problem in machine and deep learning research. This is very much the case for textual NLP datasets in English and other major world languages. For the Bangla language, the situation is even worse and the number of large datasets for NLP research is practically nil. 
There is little research undertaken on the Bangla text classification challenge applied to small datasets and limited article categories. 
The situation is evident from Table~\ref{tab.relatedResearch} that compares Potrika with the existing related datasets and shows that (apart from Potrika) the existing datasets for newspaper classification in the Bangla language are limited in their size and other attributes. 

The largest dataset after the Potrika dataset (see Table~\ref{tab.relatedResearch}) is the BARD \cite{alam2018bard} dataset that provides news articles alone without giving the headings, publication dates, and the newspaper sources of the articles. Also, the BARD \cite{alam2018bard} dataset is unbalanced. It contains around 240K articles related to the State or National category, whereas the categories Economy, International, Entertainment, and Sports have 19K, 32K, 31K, and 50K news articles, respectively. An unbalanced dataset is not suitable for news article classification. The other datasets mentioned in the table, Hossan~\cite{zakir40k} and Shahin~\cite{shahin2020classification}, have a much smaller number of news articles and these are not suitable for deep learning algorithms. Moreover, the BARD and Hossan datasets were provided in 2018 and therefore these are relatively old. 
The comparison given in the table makes it evident that there is a scarcity of balanced and comprehensive datasets and NLP research for Bangla news articles.

Potrika comprises two datasets, a raw dataset, which is described in Section~\ref{sec:description.raw} and a balanced dataset described in Section~\ref{sec:description.balanced}.

\begin{figure}[h]
	\begin{center}
		\includegraphics[scale=0.40]{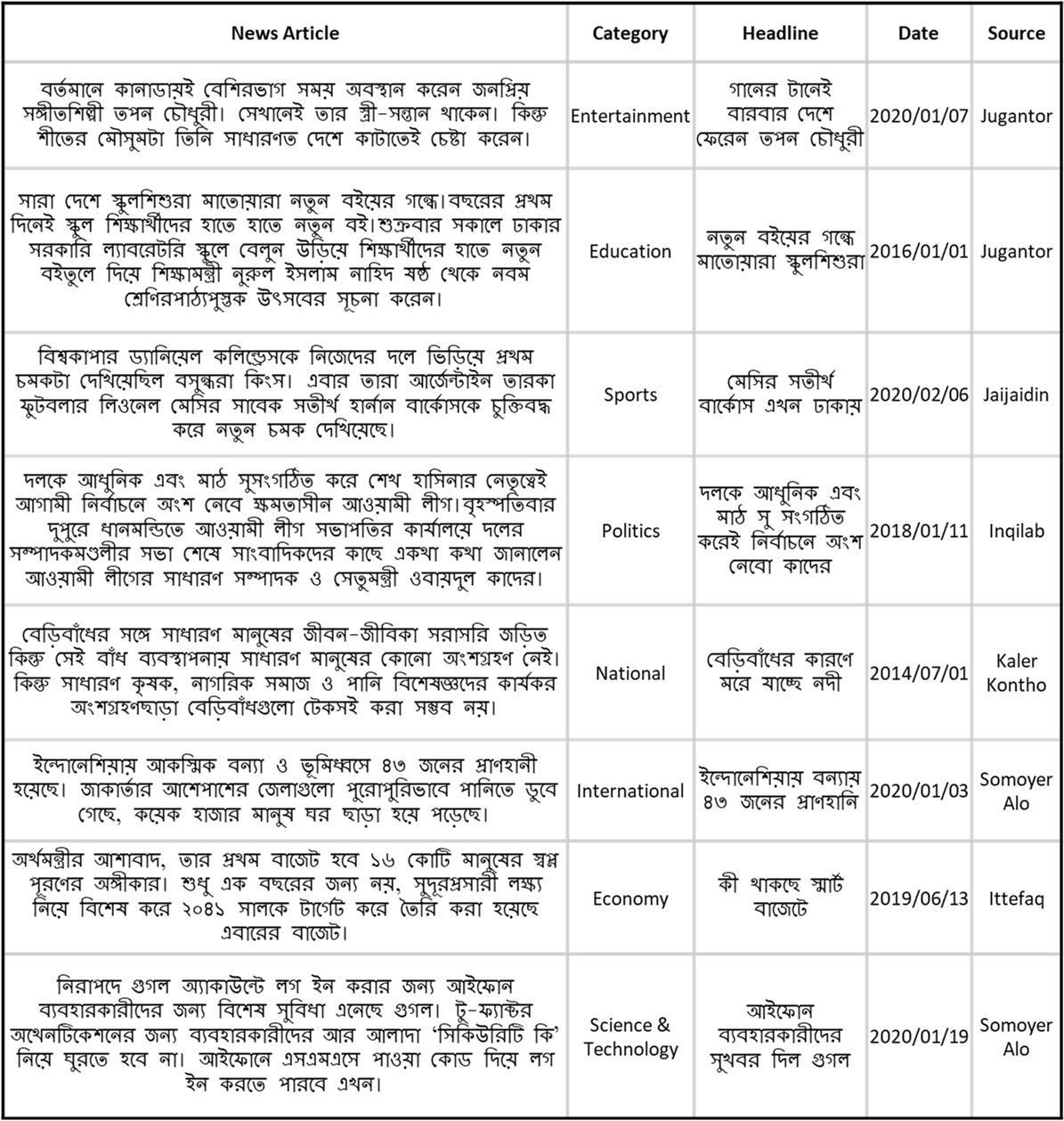}
		\caption{Potrika: News Article Examples for Each News Category}
		\label{fig.dataset_example}
	\end{center}
\end{figure}

\subsection{Potrika: The Raw Dataset}\label{sec:description.raw}

The dataset contains 665K articles divided into eight distinct categories (National, Sports, International, Entertainment, Economy, Education, Politics, and Science \& Technology). We have collected five attributes for each document that are News Article, Category, Headline, Publication Date, and Newspaper Source. The dataset can be used in a variety of natural language processing tasks such as single-label text classification, text summarization, named entity recognition, word embedding model generation, news analysis task, machine translation, and so on.
Figure \ref{fig.dataset_example} shows one example of news article for each of the eight categories with the respective four other attributes Category, Headline, Publication Date, and Newspaper Source.  

In the raw dataset folder, the news articles are organised in eight different directories respective to each of the eight news categories. Each directory contains eight CSV files for the six newspapers (one CSV file for each newspaper except Jugantor and Ittefaq, which have two files each due to their website structure). Each CSV file has five columns, one for each of the five attributes that we have collected for each news article. These five attributes have already been mentioned. After web scraping, the data is maintained in its raw form with no cleaning, stemming, or other pre-processing. Duplicate articles were removed from the dataset. Some English symbols, punctuations, arithmetic, and special characters can be found throughout the articles.  

\begin{figure}[h]
	\begin{center}
		\includegraphics[scale=0.52]{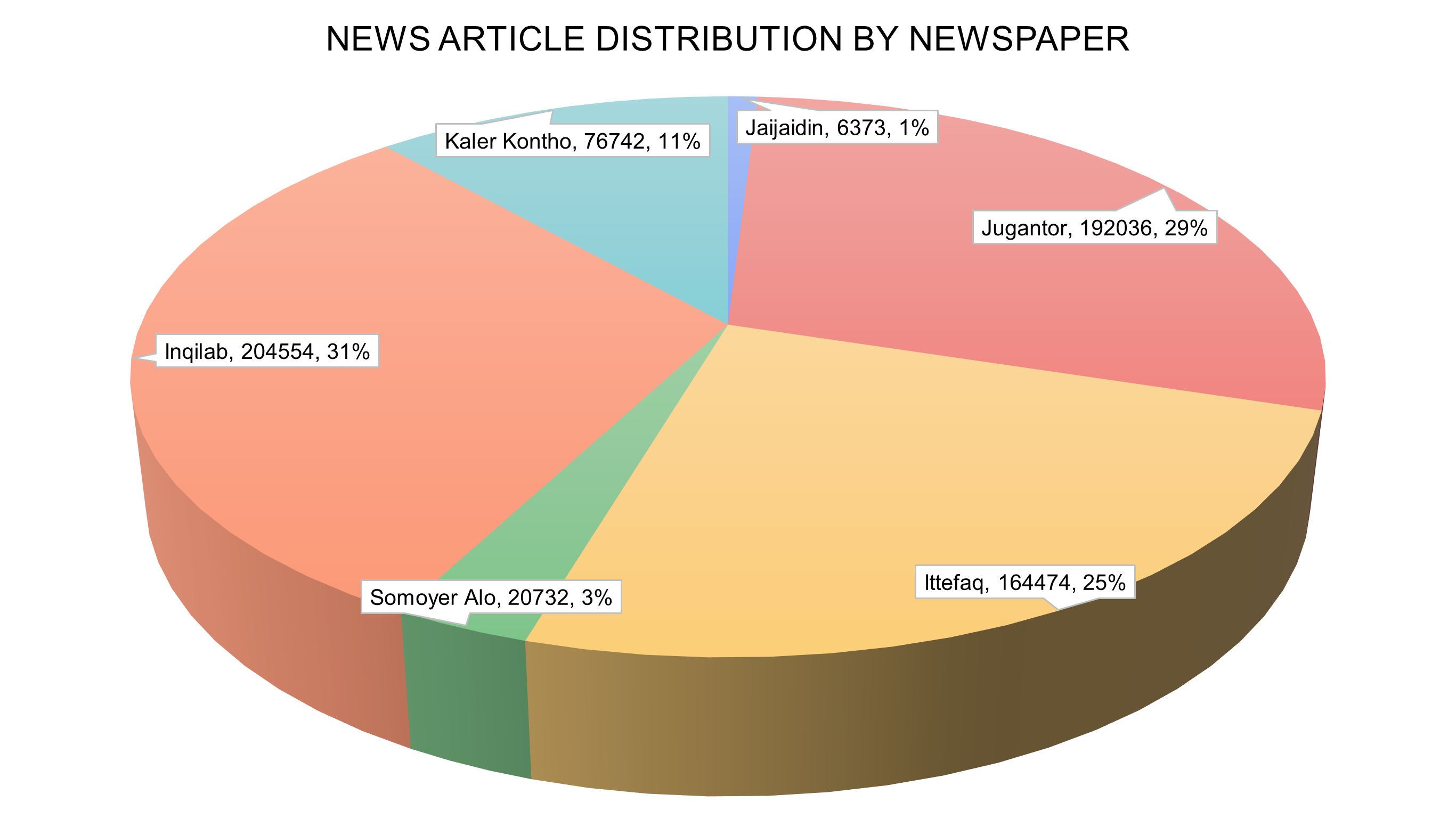}
		\caption{Potrika: News Article Distribution by Newspaper}
		\label{fig.Newspaper_based}
	\end{center}
\end{figure}

Figure~\ref{fig.Newspaper_based} depicts the number and percentage of the news articles for each newspaper. Most of the news articles (around 85\%) belong to the top three newspapers: Inqilab, Jugantor, and Ittefaq. The newspapers Kaler Kontho, Somoyer Alo, and Jaijaidin have contributed to 11\%, 3\% and 1\% of the news articles, respectively. Additional information about the distribution of news articles across the six newspapers and eight news categories is shown in Table~\ref{tab.artcledistribution}. The table shows that most of the news articles came from the Inqilab newspaper. Note that the National category has the highest number of articles among all the eight categories. The Kaler Kontho newspaper archive has zero articles in the Entertainment category.

\begin{table}[h]
	\centering
	\caption{Distribution of News Articles across Six Newspapers and Eight News Categories}
	\label{tab.artcledistribution}
	\begin{tabular}{lrrrrrrr} 
		\hline
		& Inqilab & Jugantor & Ittefaq & \begin{tabular}[c]{@{}c@{}}Kaler \\ Kontho\end{tabular} & \begin{tabular}[c]{@{}c@{}}Somoyer\\ Alo\end{tabular} & Jaijaidin & \textbf{Total}            \\ 
		\hline
		National      & 103602  & 82322    & 80388   & 18003        & 6365        & 4424      & \textbf{295104}                      \\
		\hline
		International & 47026   & 30352    & 32618   & 7            & 4026        & 545       & \textbf{114574}                      \\
		\hline
		Sports        & 26197   & 38550    & 18683   & 17618        & 3492        & 499       & \textbf{105039}                      \\
		\hline
		Entertainment & 14898   & 8134     & 11414   & 0            & 2906        & 333       & \textbf{37685}                       \\
		\hline
		Economy       & 11156   & 6084     & 3718    & 15306        & 2475        & 47        & \textbf{38786}                       \\
		\hline
		Education     & 663     & 4838     & 3225    & 12704        & 426         & 171       & \textbf{22027}                       \\
		\hline
		ScienceTech   & 544     & 5993     & 5179    & 11476        & 396         & 53        & \textbf{23641}                       \\
		\hline
		Politics      & 468     & 15733    & 9248    & 1628         & 646         & 301       & \textbf{28024}                       \\
		\hline
		\hline
		\textbf{Total}         & \textbf{204,554}  & \textbf{192,006}   & \textbf{164,473}  & \textbf{76,742}        & \textbf{20,732}       & \textbf{6,373}      & \textbf{664,880}                      \\
		\hline
	\end{tabular}
\end{table}

Table~\ref{tab.numbeASW} lists the number of news articles, sentences, and words for each of the eights categories in the Potrika raw dataset. There are a total of about 665K news articles that contain 12.57 million sentences and 185.51 million words. The data is presented in the table the descending order of the number of news article in each category. The National category has the most articles (295K) in the dataset, followed by the Sports and International categories, 105K and 114K articles, respectively. Education and Science \& Technology categories have the lowest number of articles compared to the other categories.

\begin{table}[h]
	\centering
	\caption{Potrika Raw Dataset: Number of News Articles, Sentence, and Words per Category}
	\label{tab.numbeASW}
	\begin{tabular}{lrrr} 
		\hline
		Category      & ~~~~~Articles (K) & ~~~~~Sentence (M) & ~~~~~Words (M)  \\ 
		\hline
		National      & 295                  & 5.46                  & 87.34            \\ 
		\hline
		International & 114                  & 1.80                 & 26.34            \\ 
		\hline
		Sports        & 105                  & 2.21                  & 29.81             \\ 
		\hline
		Economy       & 39                   & 0.71                 & 11.65              \\ 
		\hline
		Entertainment & 38                   & 0.62                 & 8.01            \\ 
		\hline
		Politics      & 28                   & 0.53                 & 8.4              \\ 
		\hline
		Science \& Technology   & 24                   & 0.32                  & 4.8            \\ 
		\hline
		Education     & 22                   & 0.92                 & 9.16             \\ 
		\hline
		\hline
		\textbf{Total}         & \textbf{665}                  & \textbf{12.57}                & \textbf{185.51}           \\
		\hline
	\end{tabular}
\end{table}

\begin{figure}[h]
	\begin{center}
		\includegraphics[scale=0.37]{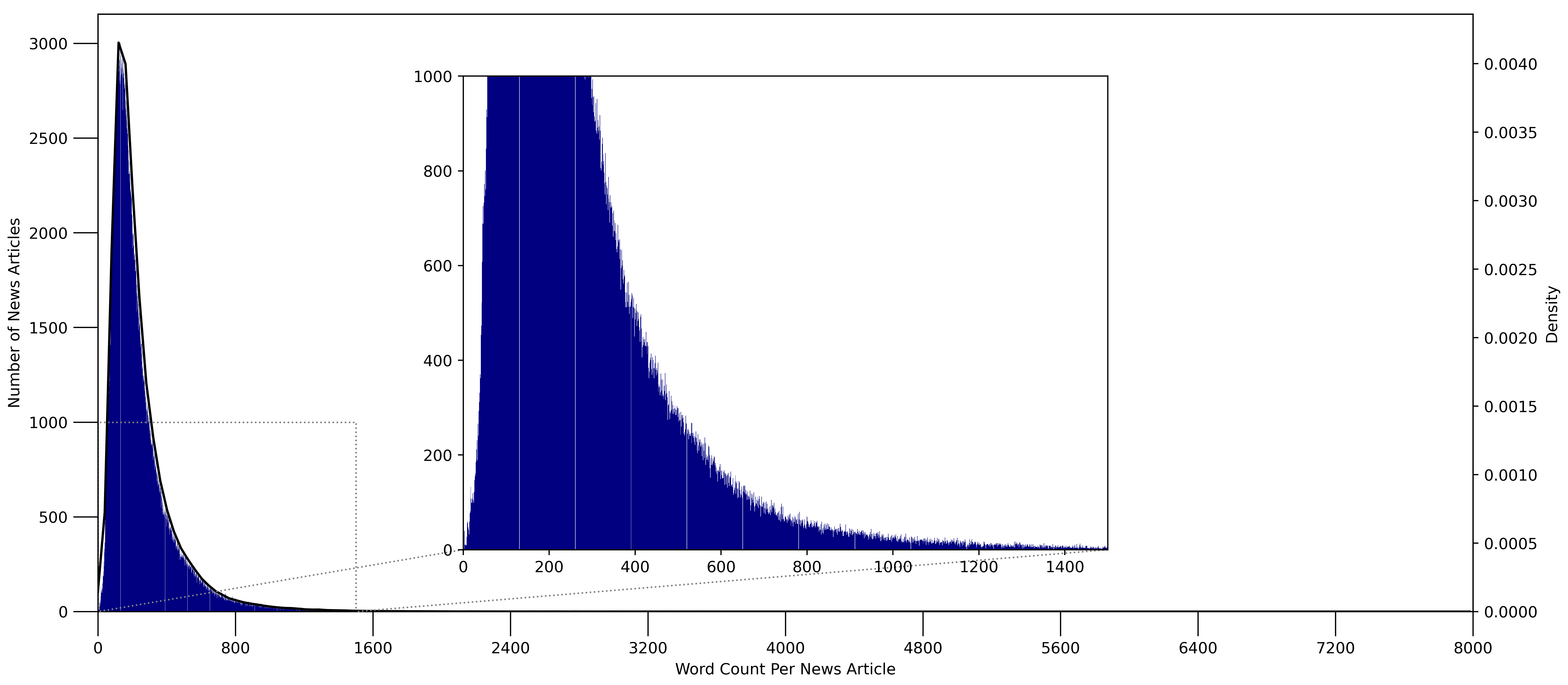}
		\caption{Number of News Articles against Word Count Per News Article}
		\label{fig.neswArticleWordCount}
	\end{center}
\end{figure}

Figure~\ref{fig.neswArticleWordCount} plots the number of news articles (the blue bars) for an increasing number of words per document (news article). The majority of news articles contain between 100 to 250 words (see the graph peaks). The number of news articles that contain over 800 words is relatively small. The maximum number of articles in any document in the data is approximately 8000 (7939 to be precise). The maximum number of articles or documents for any given number of words is around 3000. The zoomed plot inside the figure is provided for the reader's convenience. The figure also plots the density against an increasing number of words per document. See the y-axis on the right of the figure and the black plot around the blue bars. The maximum density is around 0.004.

\begin{figure}[h]
	\begin{center}
		\includegraphics[scale=0.33]{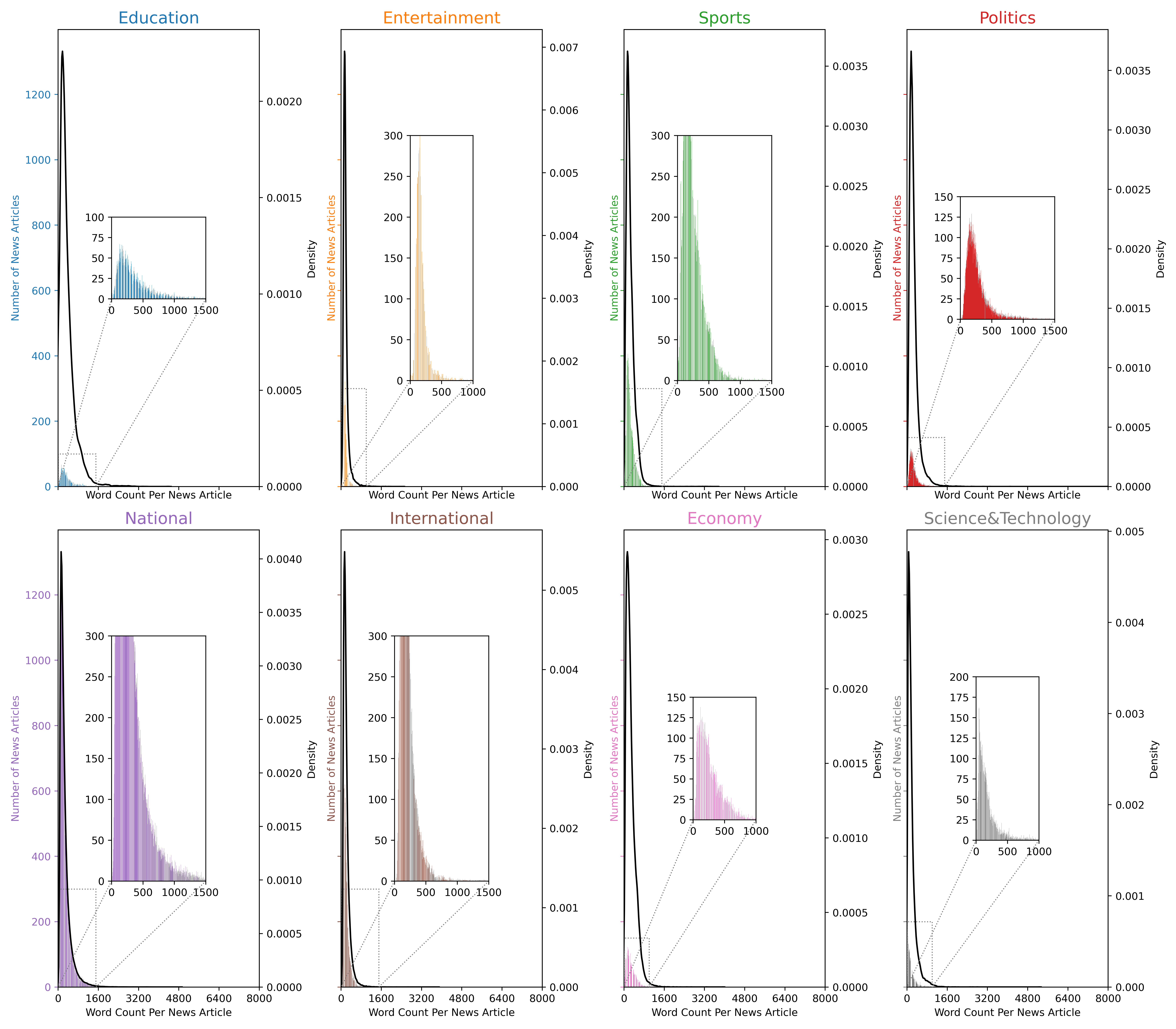}
		\caption{Density and Number of News Articles against Word Count for Each Category}
		\label{fig.Newspaper_perCategory}
	\end{center}
\end{figure}

Figure~\ref{fig.Newspaper_perCategory} plots density and the number of news articles against the word count per document for each category. The zoomed plots inside the figures are provided for the reader's convenience. Note that the National news category holds the highest number (around 1200) of news articles comprising around 300 words per article (for the peak). The International category has just over 600 news articles with this peak (the maximum number of articles) of around 250 words per article. The majority of news articles on entertainment, sports, politics, and the economy are around 500 words in length (look at the peaks in the zoomed figures). Note also that the Entertainment news category has the highest density of about 0.007 whereas Education has the least density of just over 0.002. The reason for the low density for the Education category is the fact that the maximum number of news articles is relatively low (just over 50) and are spread across the wider spectrum of the number of words per document (up to 1500 or more).

\subsection{Potrika: The Balanced Dataset}\label{sec:description.balanced}
An unbalanced dataset where the number of instances in the training dataset per class label is not balanced can produce low accuracy for a predictive classification model. In the Potrika dataset, we added a folder called ``BalancedDataset'' that contains a balanced dataset where each news category has 40,000 articles and the total number of articles for all the categories is 320,000. The balanced dataset is created using NLP augmentation techniques that will be described in Section~\ref{sec.design.balance}.

\section{Design and Methodology}\label{sec.design} 
\begin{figure}[h]
	\begin{center}
		\includegraphics[scale=0.1]{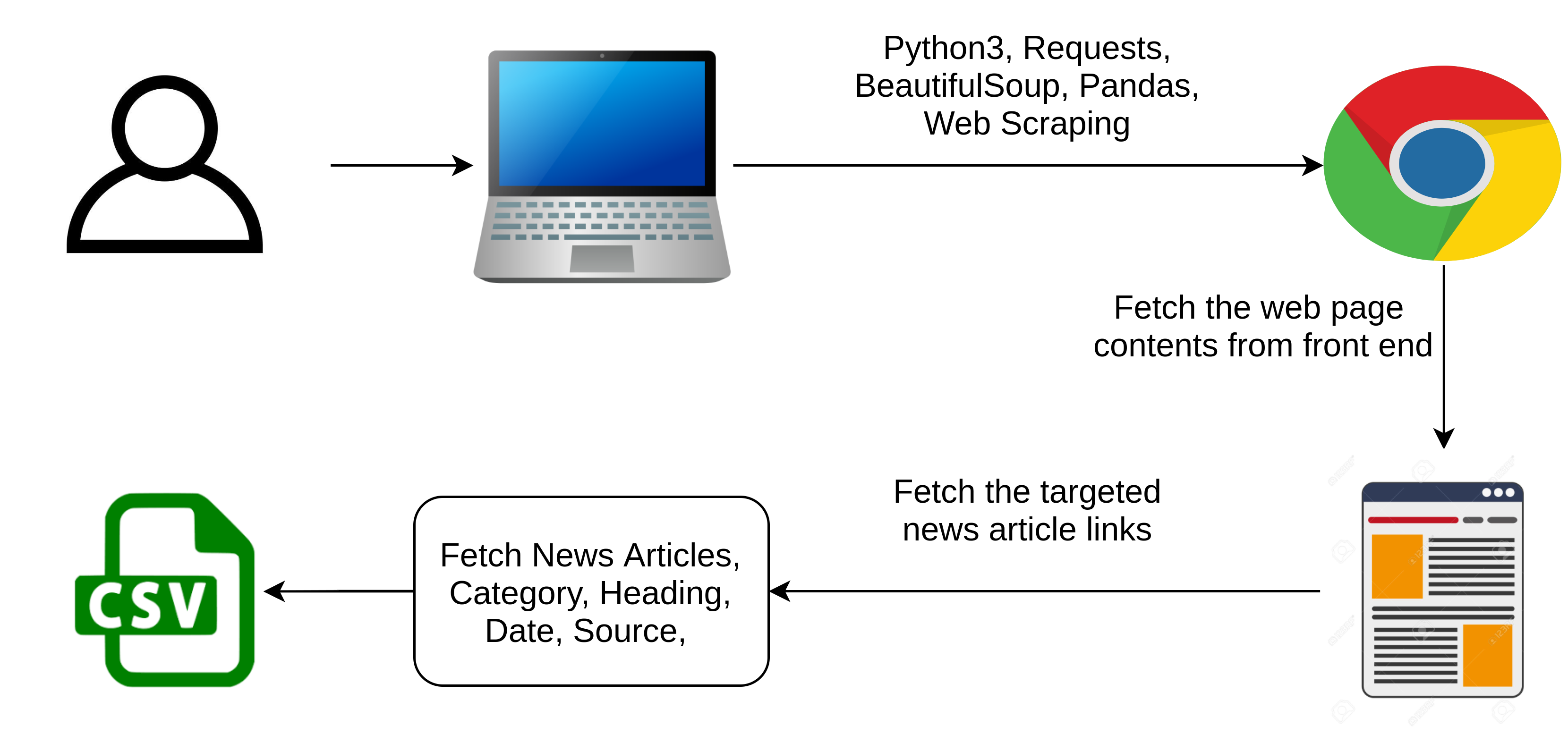}
		\caption{Dataset Collection Architecture}
		\label{scraping_process}
	\end{center}
\end{figure}

Figure~\ref{scraping_process} depicts a high-level architecture of our data collection process. The web scraping and Python tools including BeautifulSoup, Requests, Pandas, and others were used to collect the articles. Our data gathering procedure was separated into two parts. First, in algorithm~\ref{Algo.1}, article links (URL) are gathered from each newspaper archive, and then algorithm \ref{Algo.2} retrieves the news articles, Category, Heading, Publish Date, and Source from the link using the HTML and JavaScript tags. Finally, we export the data into a CSV file.

\subsection{Link Collection for News Articles} 
Initially, we assembled the links to the news from six Bangla newspaper archives using web scraping and python libraries including BeautifulSoup, Requests, Pandas, etc. There are two different versions of online archives for the newspapers Jugantor and Ittefaq; for other newspapers, there are single online archives. There are many additional links in the archive, but we deleted all the additional links and only saved the targeted links.

\begin{algorithm}[H]
	\caption{News Articles Link Collection}
	\label{Algo.1}
	\begin{algorithmic}[Algorithm 1]
		\Procedure{articlelink}{$archive$} \Comment{Newspaper archive link}
		
		\For{\texttt{<startYear to endYear>}}
		\For{\texttt{<startMonth to endMonth>}}
		\For{\texttt{<startDay to endDay>}}
		\State $url\gets archive/year/month/day$
		\State $content \gets \textit{fetch the content from the URL}$
		\State $links \gets \textit{save the link from the content}$ \Comment{News articles link}
		\EndFor
		\EndFor
		\EndFor
		\EndProcedure
	\end{algorithmic}
\end{algorithm}

\subsection{Data Collection} 
Retrieving the data from the website automatically is most challenging because an enormous amount of additional data is frequently available on the website link. Initially, we analyzed the pattern of the URL (Uniform Resource Locator) and found that all the newspapers have almost the same pattern in their online archive URLs. For example, the pattern is like ``newspaperlink/archive/year/month/day''  or ``newspaperlink/category/articleID'', etc. Based on the pattern, we separately fetch the content of the URL from different newspapers and retrieved the targeted data.

\begin{algorithm}[H]
	\caption{Data Collection}
	\label{Algo.2}
	\begin{algorithmic}[Algorithm 2]
		\Procedure{datacollection}{$ links$} 
		
		\For{\texttt{<i = 0 to total number of links>}}
		\State $link \gets links[i]$
		\State $content \gets \textit{fetch the content from the link}$
		\State $article\gets \textit{fetch article from content}$
		\State $headline\gets \textit{fetch headline from content}$
		\State $category\gets \textit{fetch category from content}$
		\State $time\gets \textit{fetch publication date from content}$
		\EndFor
		\State \texttt{save the dataset as CSV file}
		\EndProcedure
	\end{algorithmic}
\end{algorithm}

\subsection{Balancing the Dataset using NLP Data Augmentation}\label{sec.design.balance} 

NLP data augmentation techniques are most commonly used on tiny and unbalanced datasets. There are three well-known augmentation strategies. \%1) Back Translation: this approach involves translating the text material into another language and then back into the original language. This technique preserves the context of the textual data. 2) Easy Data Augmentation (EDA) consists of four methods: synonym replacement, random insertion, deletion, and text swapping. 3) NLP Albumentation: For duplicate value, sentences are shuffled or excluded. 

\begin{enumerate}
	
	\item Back Translation: this approach involves translating the text material into another language and then back into the original language. This technique preserves the context of the textual data. 
	
	\item Easy Data Augmentation (EDA) consists of four methods: synonym replacement, random insertion, deletion, and text swapping. 
	
	\item NLP Albumentation: for duplicate value, sentences are shuffled or excluded. 
	
\end{enumerate}

The Potrika dataset comprises a relatively small number of documents for five categories (Education, Economy, Science \& Technology, Politics, and Entertainment) compared to the other three categories, National, International, and Sports, which have more than 100k articles for each category. We used back translation methods on the datasets from these five categories to create a balanced dataset. We use the Google Translator API to convert Bangla text to English and then back to Bangla. The Google Translate API only accepts legitimate text with a limit of 5000 characters; otherwise, it cannot be translated. Each of the news category in the balanced dataset has 40,000 articles after performing the data augmentation approach. The balanced dataset has 320,000 news articles in total (each of the eight categories have 40,000 articles).

\section{Conclusion, Utilization, and Outlook} \label{sec.conclusion}
The Potrika dataset introduced in this paper is the largest and the most extensive dataset for news classification.
Textual datasets are in critical demand for NLP and machine learning research and development, in all human languages including English, more so in languages such as Bangla where NLP research is limited. Apart from our dataset Potrika, the existing datasets for newspaper classification in the Bangla language are limited in their size, attributes, and other properties. 

We introduced the Potrika dataset in this paper, a large single-label Bangla news article textual dataset curated for NLP research from six popular online news portals in Bangladesh. The articles were classified into eight distinct categories providing five attributes for each news article.  
The raw dataset contains over 185 million words and 12 million sentences contained in nearly 665 thousand news articles. Moreover, we created and included in Potrika an additional dataset that is balanced to facilitate those who intend to research in NLP involving a balanced dataset. 
The Potrika raw and balanced datasets can be used for text classification, text summarization, text generation, and other NLP-related research. Potrika is made available publicly and freely. 

We are working on providing applications of NLP research over the Potrika datasets, particularly on investigating performance of machine and deep learning algorithms for news classification and on the use of clustering for automatic labelling. The research is completed, is in the write-up stage and will soon be submitted for publication. Our earlier research on NLP has focussed in multiple languages on a range of problems including transportation~\cite{ahmad2022deep, Alomari2019, Alomari2021}, logistics~\cite{Suma2020, Suma2017}, healthcare~\cite{Alotaibi2020, Alomari2021a, 10.3389/frsc.2022.871171}, and urban development and governance~\cite{Alsulami2018, Alotaibi2019, Yigitcanlar2020c}. These works fall under the broader umbrella of smart cities and societies~\cite{Mehmood2020, Yigitcanlar2020e} using big data and artificial intelligence technologies~\cite{Yigitcanlar2021, Yigitcanlar2021a} where we have developed significant research outputs. We plan to extend our research on the Bangla language for these problems in the future. This will involve curating additional open datasets in the Bangla language and developing NLP research over Potrika and additional datasets.

\section*{Ethics Statement}
The reuse of article text from Bangladesh's online newspaper complies with the terms of use. All the articles were acquired with the consent of the people, groups or organizations.

\section*{Acknowledgments}
The work reported in this paper is supported by the High Performance Computing Centre (HPC Center) at King Abdulaziz University, Saudi Arabia. The computing tasks reported in this paper were performed on the Aziz supercomputer at the HPC Center, King Abdulaziz University. We would also like to express our gratitude to the authors of Jugantor, Inqilab, Jaijaidin, Kaler-kontho, Ittefaq, and Somoyer Alo, for making their news article archives publicly available.

\section*{Declaration of Competing Interest}
The authors declare that they have no known competing financial interests or personal relationships that could have appeared to influence the work reported in this paper.

\bibliographystyle{unsrt}  
\bibliography{Potrika}

\end{document}